\documentclass[letterpaper,journal]{IEEEtran}
\usepackage{amsmath,amsfonts}
\usepackage{algorithmic}
\usepackage{algorithm}
\usepackage{array}
\usepackage[caption=false,font=normalsize,labelfont=sf,textfont=sf]{subfig}
\usepackage{textcomp}
\usepackage{stfloats}
\usepackage{url}
\usepackage{verbatim}
\usepackage{graphicx}
\usepackage{cite}
\usepackage{xcolor}
\usepackage{makecell} 
\usepackage{bm}
\usepackage[export]{adjustbox}
\usepackage{tabularx}
\usepackage{multirow}
\usepackage{tikz}
\usepackage{algorithm}
\usepackage[shortlabels]{enumitem}
\usepackage{booktabs}
\usepackage{colortbl}
\usepackage{setspace}
\usepackage{wrapfig}

\definecolor{myciteblue}{RGB}{40,110,220}
\definecolor{myrefyellow}{RGB}{180,110,0}
\usepackage{pifont}
\usepackage{bbding}
\usepackage[most]{tcolorbox}
\usepackage{placeins}
\usepackage{subfig}

\usepackage{mathtools}
\usepackage{multicol}
\usepackage[font=small]{caption}
\usepackage{comment}
\usepackage{float}
\usepackage{amssymb}
\usepackage{amsthm}

\begin{document}

\title{Cognitive Action Reasoning for Proactive Robots from Human-Centered Multimodal Observations}

\author{Zhihao Gu$^{1,*}$\thanks{$^{*}$Equal contribution. Work done when Zhihao Gu was a Postdoc at NTU.}, Kechao Zhu$^{1,*}$, Yuanfeng Wu$^{1,*}$, Mohan Liu$^{1,*}$, \\ Ankit Kumar Shaw,, ChenDong Hong$^{1}$, Xuanyu Chen$^{1}$, Dengchen Mei$^{1}$, Xu Tianyi$^{1}$, Lin Wang$^{1}$\\
School of Electrical and Electronic Engineering, Nanyang Technological University (NTU)
        % <-this % stops a space
% \thanks{Manuscript received April 19, 2026; revised June 12, 2026.}
}

% \markboth{IEEE ROBOTICS AND AUTOMATION LETTERS,~Vol.~11, No.~6, June~2026}%
\markboth{Journal of \LaTeX\ Class Files,~Vol.~18, No.~9, June~2026}%
{How to Use the IEEEtran \LaTeX \ Templates}

% \IEEEpubid{0000--0000~\copyright~2026 IEEE}

\maketitle

\begin{abstract}
Robots operating in human-centered environments are typically designed to execute explicit instructions, and most robot-learning datasets likewise pair observations with task instructions or low-level actions. Although recent work has begun to explore proactive embodied assistance, existing resources target different settings and action levels, leaving real-world human-centered multimodal decision-making underexplored. We formulate this problem as \textit{Proactive Robot Action Reasoning} (\textit{ProRobo}), an upstream cognitive decision problem in which a robot must determine which action to take based on multimodal human and environmental cues without explicit action instructions. To support ProRobo, we introduce \textit{ProAction}, a real-world multimodal dataset containing 10K samples of visual observations, audio signals, and text inputs across 12 daily-life scenarios in five common scenes. 
To construct cognitively grounded high-level action supervision, we develop a two-stage human-in-the-loop pipeline that combines appraisal-guided candidate generation with Affective Theory-of-Mind-guided human refinement, explicitly incorporating contextual judgment about human states, urgency, feasibility, and potential risk into action annotation.
Based on this supervision, we benchmark representative Multimodal Large Language Models (MLLMs) and introduce \textit{MMC2Act}, a reference model that implicitly learns the mapping from multimodal observations to cognitively grounded high-level actions. Experiments across modality settings, subject-disjoint generalization, cross-dataset transfer, and human evaluation show that general-purpose MLLMs struggle with proactively reasoning high-level actions from multimodal cues, whereas training on \textit{ProAction} substantially improves performance. Finally, physical-robot experiments show that the inferred high-level actions can be grounded into executable robot behaviors.
\end{abstract}


\section{Introduction}\label{intro}
\IEEEPARstart{R}{obots} deployed in human-centered environments are predominantly \textit{reactive}: a user specifies what the robot should do, and the robot then executes the corresponding action. While this paradigm has enabled substantial progress in robot learning and embodied intelligence, it assumes that humans explicitly formulate their desired robot actions. In everyday interactions, however, humans do not always issue direct commands. A person may cough while eating, struggle to reach an object, or express excitement through speech and body behavior. Therefore, enabling robots to operate proactively requires more than multimodal perception or instruction following. Robots need a cognitive decision layer that reasons over human states and surrounding context to determine what action should be taken before downstream execution.

\begin{figure}[t]
  \centering
 \includegraphics[width=0.95\columnwidth]{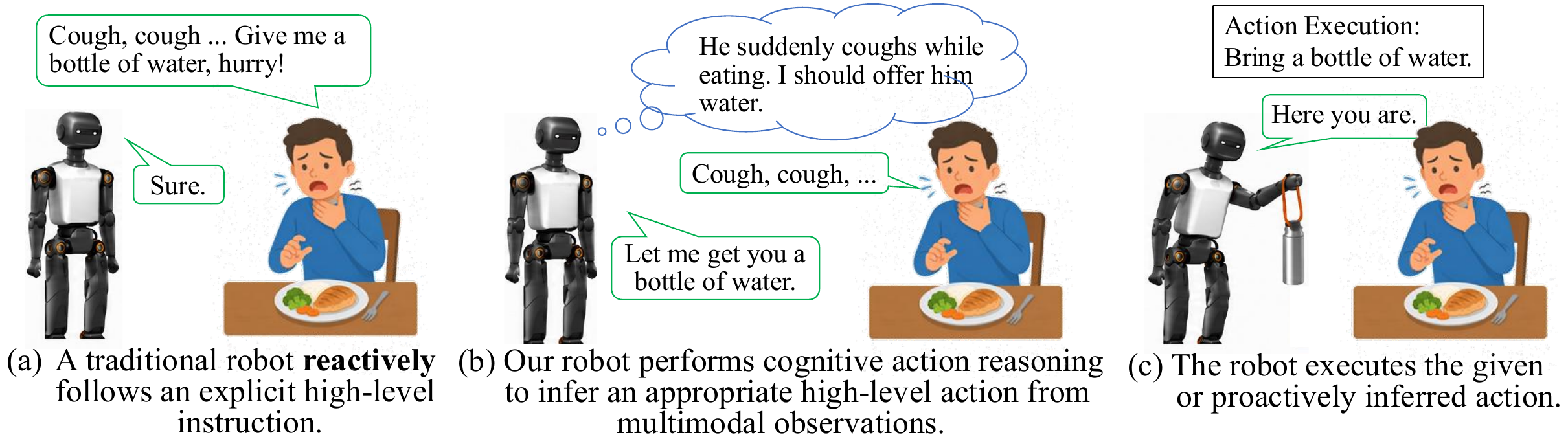}
  % \vspace{-4pt}
   \caption{\textbf{Paradigm Comparison.} (a) A conventional robot reacts to an explicit instruction specifying what it should do. (b) The robot in our setting receives only human-centered observations, such as body posture, facial expressions, and vocal tone, and proactively reasons about an appropriate high-level action based on them. (c) The inferred high-level action is then grounded into an executable robot behavior.}\label{motivation}
   \vspace{-8pt}
\end{figure}

% We study this upstream cognitive decision problem as \textit{Proactive Robot Action Reasoning (\textit{ProRobo})}. Given multimodal observations of a human and the surrounding environment, ProRobo requires a robot to infer an appropriate high-level robot action without an explicit instruction specifying what action the robot should perform.
% In this work, cognition is operationalized at the high-level action decision layer, where human states and contextual evidence are considered together with urgency, feasibility, and potential risks to determine what action should be taken before downstream execution.
% As illustrated in Fig.~\ref{motivation}, a person may suddenly cough while eating. From the observed human behavior, auditory cues, and surrounding context, the robot must infer an appropriate high-level action, such as bringing a bottle of water,  without being explicitly instructed what to do. Thus, ProRobo requires not only understanding human-centered multimodal cues, but also translating them into contextually appropriate, helpful, and feasible actions.
We study this upstream cognitive decision problem as \textit{Proactive Robot Action Reasoning (ProRobo)}, where a robot infers an appropriate high-level action from human-centered multimodal observations without explicit action instructions. In this work, cognition is operationalized at the high-level action decision layer, where human states and contextual evidence are considered together with urgency, feasibility, and potential risks before downstream execution. As illustrated in Fig.~\ref{motivation}, a person may suddenly cough while eating, requiring the robot to interpret the observed behavior, auditory cues, and surrounding context and respond appropriately, such as by bringing a bottle of water. Thus, ProRobo goes beyond multimodal understanding by requiring contextually appropriate, helpful, and feasible action decisions.

Recent advances in Multimodal Large Language Models (MLLMs) provide a promising foundation for this capability, owing to their ability to jointly process vision, audio, and text \cite{zhu2023minigpt,guo2025stimuvar,lei2024large,gu2026learning,gu2026limode}. Meanwhile, related studies have begun to explore proactive embodied assistance from contextual cues, including unspoken-need inference, empathetic task planning, audio-visual planning, and contextual manipulation~\cite{ding2024atom,chen2025empathyagent,PEAP2026,Roboomni2026,ProactiveDialog}. These efforts show growing interest in proactive embodied intelligence, but target different data settings and action levels. In particular, existing resources focus on synthetic language interactions, empathy-oriented assistance, scene-level audio-visual planning, or manipulation-oriented low-level execution. Consequently, the upstream cognitive decision problem of determining \emph{what a robot should do} from real-world human-centered multimodal observations remains underexplored.
% Consequently, the upstream problem of determining \emph{what a robot should do} from real-world human-centered multimodal observations remains underexplored.
% explored human intention understanding, affective Theory-of-Mind reasoning, and empathetic action generation \cite{ding2024atom,chen2025empathyagent,gandhi2023understanding,jin2024mmtom,zhang2025mindpower}. For instance, AToM-Bot \cite{ding2024atom} investigates how embodied agents infer human intentions from visual observations, while EmpathyAgent \cite{chen2025empathyagent} studies empathetic action generation from video and language descriptions. Nevertheless, existing datasets remain limited. Many of them are based on text-only or video-text inputs, rely on simulated data, or are not designed to associate \textbf{real-world multimodal human-centered observations with robot-oriented high-level actions}. Audio is particularly underrepresented, although it can provide complementary evidence for determining what a robot should respond.

\begin{figure}[t]
  \centering
 \includegraphics[width=0.9\linewidth]{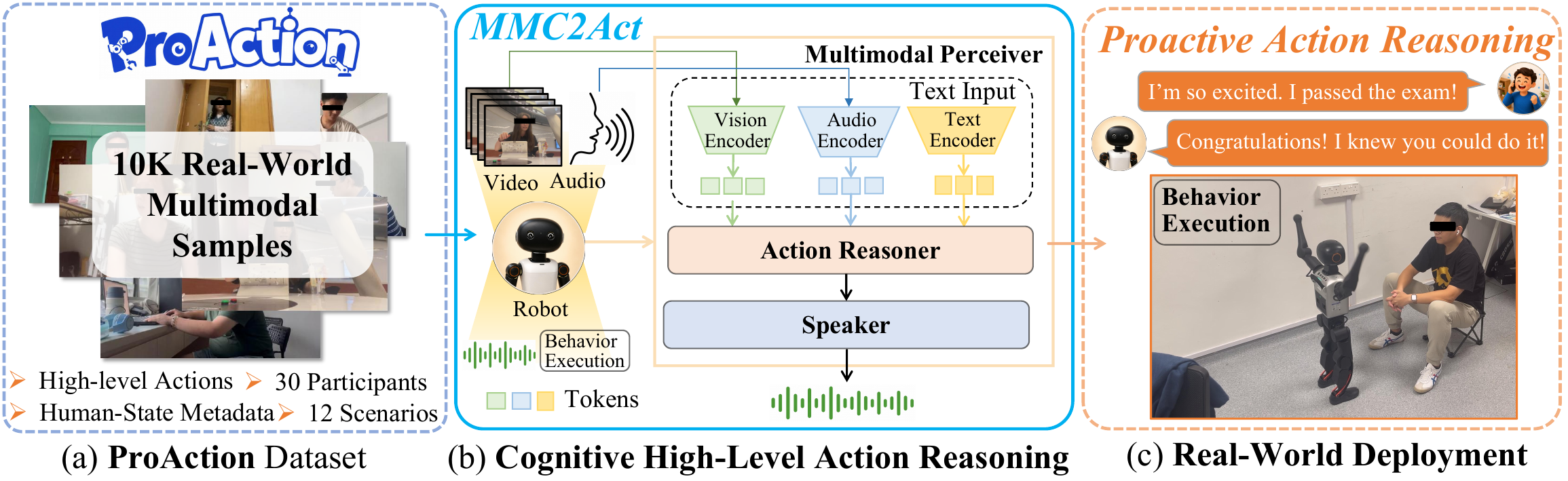}
  % \vspace{-4pt}
   \caption{\textbf{Overview}. We introduce \textit{ProAction} to support ProRobo and use a validation model to examine the utility of ProAction. Finally, we conduct \textit{real-world robot deployment} to evaluate whether the inferred high-level actions can be grounded into executable robot behaviors.}\label{framework}
   \vspace{-8pt}
\end{figure}

To support this problem, we introduce \textit{ProAction} (Fig.~\ref{framework} (a)), a real-world multimodal dataset designed to provide cognitively grounded supervision for ProRobo. The dataset comprises 10K samples of visual observations, audio signals, and text inputs collected across 12 daily-life scenarios in five common scenes. Unlike datasets centered on synthetic interactions, scene-level assistance, or trajectory-level manipulation, \textit{ProAction} associates human behaviors, states, and surrounding context with human-aligned high-level robot actions.
To obtain reliable and cognitively grounded action annotations, we develop a two-stage human-in-the-loop pipeline. In Stage I, appraisal-based criteria guide candidate action generation by considering urgency, helpfulness, feasibility, and potential risk. In Stage II, annotators refine these candidates under Affective Theory-of-Mind guidance by considering the person's states. Human-state/need interpretations are retained as auxiliary annotations for analysis and future research.
% To support this problem, we introduce \textit{ProAction} (Fig.~\ref{framework} (a)), a real-world multimodal dataset. 
% \textit{ProAction} is designed to provide cognitively grounded supervision for this decision process. Rather than directly assigning actions to observed cues, its annotation pipeline explicitly incorporates appraisal-based judgment and Affective Theory-of-Mind reasoning to derive human-aligned high-level actions from multimodal observations. The dataset comprises 10K samples of visual observations, audio signals, and text inputs collected across 12 daily-life scenarios in five common scenes.
% Unlike datasets centered on synthetic interactions, scene-level assistance, or trajectory-level manipulation, ProAction specifically associates human behaviors, states, and surrounding context with the upstream high-level action decision. 
% To obtain reliable annotation, we develop a two-stage human-in-the-loop annotation pipeline. 
% In Stage I, appraisal-based criteria guide the generation of candidate actions by considering their urgency, helpfulness, feasibility, and potential risk. In Stage II, annotators refine these candidates under Affective Theory-of-Mind guidance by considering the person's emotional and physical states. Human-state/need interpretations are further retained as auxiliary annotations for analysis and future research.

To examine whether ProAction provides effective supervision for ProRobo, we benchmark representative MLLMs and introduce \textit{MMC2Act}, a reference model in Fig.~\ref{framework} (b), under different modality and generalization settings. MMC2Act is only intended to evaluate the learnability and utility of ProAction.
% rather than introduce a highly specialized model architecture.
Experimental results show that general-purpose MLLMs still struggle to translate multimodal human-centered observations into appropriate high-level robot actions, whereas training on ProAction substantially improves performance. Finally, physical-robot experiments in Fig.~\ref{framework} (c) demonstrate that the inferred high-level actions can be grounded into executable robot behaviors, connecting upstream action reasoning with downstream execution.
To sum up, the contributions of this work are as follows:
\begin{itemize}
    \item We formulate \textit{Proactive Robot Action Reasoning} as a cognitive high-level decision problem for inferring appropriate robot actions from human-centered multimodal observations without explicit action instructions, and introduce ProAction, a 10K-scale multimodal dataset.
    % spanning 12 daily-life scenarios in five common scenes.
    % We formulate \textit{Proactive Robot Action Reasoning} as an upstream cognitive decision problem for inferring high-level actions from human-centered multimodal observations without explicit instructions, and introduce \textit{ProAction}, a 10K-scale real-world dataset spanning 12 daily-life scenarios in five common scenes.
    
    \item We develop a cognitively grounded annotation framework that explicitly incorporates appraisal-based judgment and Affective Theory-of-Mind reasoning to obtain human-aligned high-level robot actions.
    % We design a \textit{two-stage human-in-the-loop annotation pipeline} guided by appraisal-based criteria and Affective Theory-of-Mind reasoning to obtain feasible and human-aligned high-level robot-action annotations.
    % producing reliable and human-aligned high-level robot-action annotations.
    
    \item We establish a \textit{comprehensive benchmark} of representative MLLMs and the \textit{MMC2Act} reference model to assess the learnability and utility of ProAction across different modalities and generalization settings.
    % establish a \textbf{comprehensive benchmark} of representative MLLMs and the \textit{MMC2Act} baseline that assesses the learnability and utility of ProAction across different modality and generalization settings.
    
    \item We conduct \textit{physical-robot experiments} and validate that the inferred high-level actions can be grounded into executable robot behaviors.
\end{itemize}

% ------------------------------------------------------
\section{Related Works}\label{sec:formatting}
\noindent\textbf{Datasets for Robot Learning and ProRobo.}
Recent advances in robot learning have been supported by large-scale datasets for embodied perception, planning, and task execution~\cite{guo2025stimuvar,cheng2024emotion,gu2026learning,lian2025affectgpt,openx2023rtx,li2023behavior}. 
Most of these datasets condition robot behavior on explicitly specified language instructions, goals, or demonstrations. More recent resources have begun to study proactive assistance from implicit or contextual cues. AToM-Bot~\cite{ding2024atom} infers unspoken human needs from human states and visual observations and generates feasible embodied tasks, while EmpathyAgent~\cite{chen2025empathyagent} focuses on empathy-oriented task planning from multimodal context. ProactiveDialog~\cite{ProactiveDialog} constructs synthetic multi-party dialogues for proactive task inference from implicit requests. PEAP~\cite{PEAP2026} studies proactive embodied action planning through joint visual and auditory scene perception without explicit human instructions, whereas RoboOmni~\cite{Roboomni2026} introduces OmniAction for proactive robot manipulation from spoken dialogue, environmental sounds, and visual cues. These efforts demonstrate growing interest in proactive embodied intelligence, but differ in data grounding and action abstraction. Existing resources primarily focus on synthetic language interactions, empathy-specific assistance, scene-level audio-visual planning, or manipulation-oriented execution. In contrast, \textit{ProAction targets the upstream high-level action decision from real-world human-centered multimodal observations, associating human behaviors, states, and surrounding context with human-aligned robot actions.} Tab.~\ref{compar} summarizes the comparison between related datasets.
% Benchmarks such as EmbodiedBench~\cite{yang2025embodiedbench} further evaluate embodied agents across diverse capabilities. However, most existing robot-learning datasets assume that a task or target action is explicitly specified through language instructions, demonstrations, or goal descriptions. Another line of work studies social reasoning, human intention understanding, and assistive or empathetic behavior~\cite{rozo2016learning,puig2020watch,zhang2023building,gandhi2023understanding,jin2024mmtom,chen2025empathyagent,zhang2025mindpower}. For example, EmpathyAgent~\cite{chen2025empathyagent} benchmarks empathetic action generation from video and language, while AToM-Bot~\cite{ding2024atom} studies embodied assistance from real-world human-centered observations. Despite this progress, existing datasets remain limited for studying ProRobo, either because they lack real-world data, underrepresent audio, or are not designed to associate human-centered context with robot-oriented high-level actions. 
% In contrast, ProAction provides \textit{10K real-world multimodal samples} paired with human-aligned high-level actions and human-state/need interpretations as auxiliary annotations.
% It specifically targets the upstream decision problem of determining \textit{what a robot should do when no explicit instruction is provided}. Tab.~\ref{compar} summarizes the comparison.

\noindent \textbf{Cognitive Modeling for Human-Aligned Action Annotation.}
Annotating high-level robot actions from human-centered observations is inherently subjective, as an appropriate response depends on human states, contextual constraints, urgency, social norms, and potential risks. Theory of Mind~\cite{premack1978does} models reasoning about others' beliefs, goals, and intentions, while Affective Theory of Mind (AToM)~\cite{cucciniello2023mind} further emphasizes emotional and physical states. Prior work has leveraged such reasoning for social understanding and proactive assistance~\cite{ding2024atom,chen2025empathyagent,ProactiveDialog,PEAP2026}. In ProAction, \textit{these cognitive principles are used not as prediction targets, but as guidance for constructing human-aligned robot-action annotations.} Appraisal-based criteria guide candidate generation toward timely, helpful, feasible, and risk-aware actions, while AToM guides human refinement based on the person's emotional and physical states.
% Cognitive modeling provides structured principles for interpreting human states and contextual evidence before selecting an appropriate response. Theory of Mind models reasoning about others' beliefs, goals, and intentions, while Affective Theory of Mind further emphasizes emotional and physical states.
% Annotating high-level robot actions from human-centered observations is inherently subjective, because an appropriate response depends not only on the physical scene, but also on the person's observable state, urgency, contextual constraints, social norms, and potential risks. Theory of Mind~\cite{premack1978does} models reasoning about others' beliefs, goals, and intentions, while Affective Theory of Mind (AToM)~\cite{cucciniello2023mind} further emphasizes emotional and physical states. Prior work has leveraged such human-state reasoning for social understanding and proactive assistance~\cite{ding2024atom,chen2025empathyagent,ProactiveDialog,PEAP2026}. In ProAction, however, these concepts are not treated as the primary prediction target. Instead, we use them as guidance for constructing human-aligned robot-action annotations. Specifically, appraisal-based criteria guide candidate generation toward timely, helpful, feasible, and risk-aware actions, while AToM encourages annotators to consider the person's emotional and physical conditions when reviewing and refining these candidates. This provides structured supervision for mapping multimodal human-centered observations directly to appropriate high-level robot actions. 

\noindent\textbf{Proactive Robot Reasoning and Embodied Planning.}
Beyond dataset construction, recent studies have developed system-level approaches for proactive embodied agents~\cite{girdhar2023imagebind,Chen_2024_CVPR,yuan2024robopoint}. ProAct~\cite{zhang2026proact} presents a dual-system framework for proactive embodied social interaction, while event-driven assistive manipulation~\cite{liu2026eventdriven} combines event-level reasoning with grounded planning for timely robot intervention. Active perception and early action recognition~\cite{li2026actsenseact,cao2026sasi} are also relevant, as proactive agents may need to gather additional evidence or recognize evolving human activities before deciding whether and how to act. They primarily address system-level planning, perception, sensing, or downstream execution. RoboOmni~\cite{Roboomni2026} studies proactive robot manipulation by grounding multimodal cues into manipulation trajectories. 
These works mainly address system-level planning or downstream execution, whereas ProAction targets the preceding high-level action decision from human-centered multimodal observations.

\begin{table*}[t]
\centering
 \footnotesize
\caption{\textbf{Dataset comparison}. ``Human Scenario Recording'' denotes data captured from participants performing the target scenarios.
% rather than from pre-existing datasets or simulated environments. 
``Synchronized Human-Scenario A/V'' denotes audio and video jointly recorded from participants within the same human-centered scenarios. }\label{compar}
% \vspace{-5pt}
\setlength{\tabcolsep}{4pt}
\resizebox{0.6\linewidth}{!}{
\begin{tabular}{lccccccccccc}
\toprule
\multirow{2}{*}{Dataset} & \multirow{2}{*}{Proactive} &
\multicolumn{3}{c}{Modality} &
\multirow{2}{*}{\shortstack{Human Scenario\\Recording}} &
\multirow{2}{*}{\shortstack{Synchronized\\Human-Scenario A/V}} &
\multirow{2}{*}{\shortstack{Human Need\\Annotation}} &
\multirow{2}{*}{\shortstack{Robot Action \\ Annotation}} &
\multirow{2}{*}{\#Samples} \\
\cmidrule(lr){3-5}
& & Text & Audio & Video & & & & \\
\midrule
BigToM~\cite{gandhi2023understanding}
& \textcolor{purple}{\XSolidBrush}
& \textcolor{green!70!black}{\ding{52}}
& \textcolor{purple}{\XSolidBrush}
& \textcolor{purple}{\XSolidBrush}
& \textcolor{purple}{\XSolidBrush}
& \textcolor{purple}{\XSolidBrush}
& \textcolor{purple}{\XSolidBrush}
& \textcolor{purple}{\XSolidBrush} & 5K \\
MMToM-QA~\cite{jin2024mmtom}
& \textcolor{purple}{\XSolidBrush}
& \textcolor{green!70!black}{\ding{52}}
& \textcolor{purple}{\XSolidBrush}
& \textcolor{green!70!black}{\ding{52}}
& \textcolor{purple}{\XSolidBrush}
& \textcolor{purple}{\XSolidBrush}
& \textcolor{purple}{\XSolidBrush}
& \textcolor{purple}{\XSolidBrush} & 225 \\
AToM-Bot~\cite{ding2024atom}
& \textcolor{green!70!black}{\ding{52}}
& \textcolor{purple}{\XSolidBrush}
& \textcolor{purple}{\XSolidBrush}
& \textcolor{green!70!black}{\ding{52}}
& \textcolor{purple}{\XSolidBrush}
& \textcolor{purple}{\XSolidBrush}
& \textcolor{green!70!black}{\ding{52}}
& \textcolor{green!70!black}{\ding{52}} & - \\
MindPower~\cite{zhang2025mindpower}
& \textcolor{purple}{\XSolidBrush}
& \textcolor{green!70!black}{\ding{52}}
& \textcolor{purple}{\XSolidBrush}
& \textcolor{green!70!black}{\ding{52}}
& \textcolor{purple}{\XSolidBrush}
& \textcolor{purple}{\XSolidBrush}
& \textcolor{purple}{\XSolidBrush}
& \textcolor{green!70!black}{\ding{52}} & 590 \\
EmpathyAgent~\cite{chen2025empathyagent}
& \textcolor{green!70!black}{\ding{52}}
& \textcolor{green!70!black}{\ding{52}}
& \textcolor{purple}{\XSolidBrush}
& \textcolor{green!70!black}{\ding{52}}
& \textcolor{purple}{\XSolidBrush}
& \textcolor{purple}{\XSolidBrush}
& \textcolor{purple}{\XSolidBrush}
& \textcolor{green!70!black}{\ding{52}} & 10K \\
ProactiveDialog~\cite{ProactiveDialog}
& \textcolor{green!70!black}{\ding{52}}
& \textcolor{green!70!black}{\ding{52}}
& \textcolor{purple}{\XSolidBrush}
& \textcolor{purple}{\XSolidBrush}
& \textcolor{purple}{\XSolidBrush}
& \textcolor{purple}{\XSolidBrush}
& \textcolor{purple}{\XSolidBrush}
& \textcolor{green!70!black}{\ding{52}} & 10K \\
PEAP~\cite{PEAP2026}
& \textcolor{green!70!black}{\ding{52}}
& \textcolor{green!70!black}{\ding{52}}
& \textcolor{green!70!black}{\ding{52}}
& \textcolor{green!70!black}{\ding{52}}
& \textcolor{purple}{\XSolidBrush}
& \textcolor{purple}{\XSolidBrush}
& \textcolor{purple}{\XSolidBrush}
& \textcolor{green!70!black}{\ding{52}} & 19.9K \\
\midrule
\textbf{ProAction (Ours)}
& \textcolor{green!70!black}{\ding{52}}
& \textcolor{green!70!black}{\ding{52}}
& \textcolor{green!70!black}{\ding{52}}
& \textcolor{green!70!black}{\ding{52}}
& \textcolor{green!70!black}{\ding{52}}
& \textcolor{green!70!black}{\ding{52}}
& \textcolor{green!70!black}{\ding{52}}
& \textcolor{green!70!black}{\ding{52}} & 10.1K \\
\bottomrule
\end{tabular}}
\vspace{-10pt}
\end{table*}

% \section{Methodology}
% Proactive robots must reason about suitable actions without receiving explicit commands. Yet existing datasets lack real-world audio-visual grounding and human-aligned high-level action annotation. To bridge this gap, we introduce \textbf{ProAction}, a real-world dataset constructed for inferring high-level robot actions from human-centered multimodal cues, such as body posture, facial expression, vocal tone, and surroundings, \textit{which is significantly different from manipulation datasets}.

\section{Methodology}
ProRobo requires a robot to determine an appropriate high-level action from human-centered multimodal observations without explicit action instructions. We operationalize cognition at this high-level decision layer through cognitively grounded action supervision, where appraisal-based judgment and Affective Theory-of-Mind guidance are explicitly used during annotation to derive human-aligned high-level actions.
To support this problem, we construct ProAction, a real-world dataset containing audio $A$, video $V$, and textual input $T$, together with human-validated high-level robot-action annotations $\bar{Y}$. $T$ is a task-level prompt that defines the model's role and expected output format, rather than specifying the particular action to perform. We further introduce MMC2Act, a multimodal validation baseline $f:(A,V,T)\rightarrow Y$, where $Y$ denotes the inferred high-level action, to examine whether ProAction provides learnable supervision for ProRobo. Human-state/need interpretations $Z$ are retained only as auxiliary semantic annotations for analysis and future research. Thus, cognitive principles are explicitly incorporated into the construction of high-level action supervision, while MMC2Act implicitly learns the resulting observation-to-action mapping.
% ProRobo requires a robot to determine an appropriate high-level action from human-centered multimodal observations without explicit action instructions. We operationalize cognition at this high-level decision layer through cognitively grounded action supervision, where appraisal-based judgment and Affective Theory-of-Mind guidance are explicitly used during annotation to derive human-aligned high-level actions.
% To support this problem, we construct ProAction, a real-world dataset containing audio $A$, video $V$, and textual input $T$, together with human-validated high-level robot-action annotations $\bar{Y}$. $T$ is a task-level prompt that defines the model's role and expected output format, rather than specifying the particular action to perform. We further introduce MMC2Act, a multimodal validation baseline $f:(A,V,T)\rightarrow Y$, where $Y$ denotes the inferred high-level action, to examine whether ProAction provides learnable supervision for ProRobo. Rather than implementing an explicit cognitive reasoning chain, MMC2Act implicitly learns the observation-to-action mapping encoded in these high-level action annotations. Human-state/need interpretations $Z$ are retained only as auxiliary semantic annotations for analysis and future research. 
% Thus, cognition is explicit in the construction of high-level action supervision, while MMC2Act implicitly learns the resulting observation-to-action mapping.

\subsection{ProAction: A Real-World Multimodal Dataset}\label{prodataset}
\begin{figure*}
  \begin{center}
  \centerline{\includegraphics[width=0.85\linewidth]{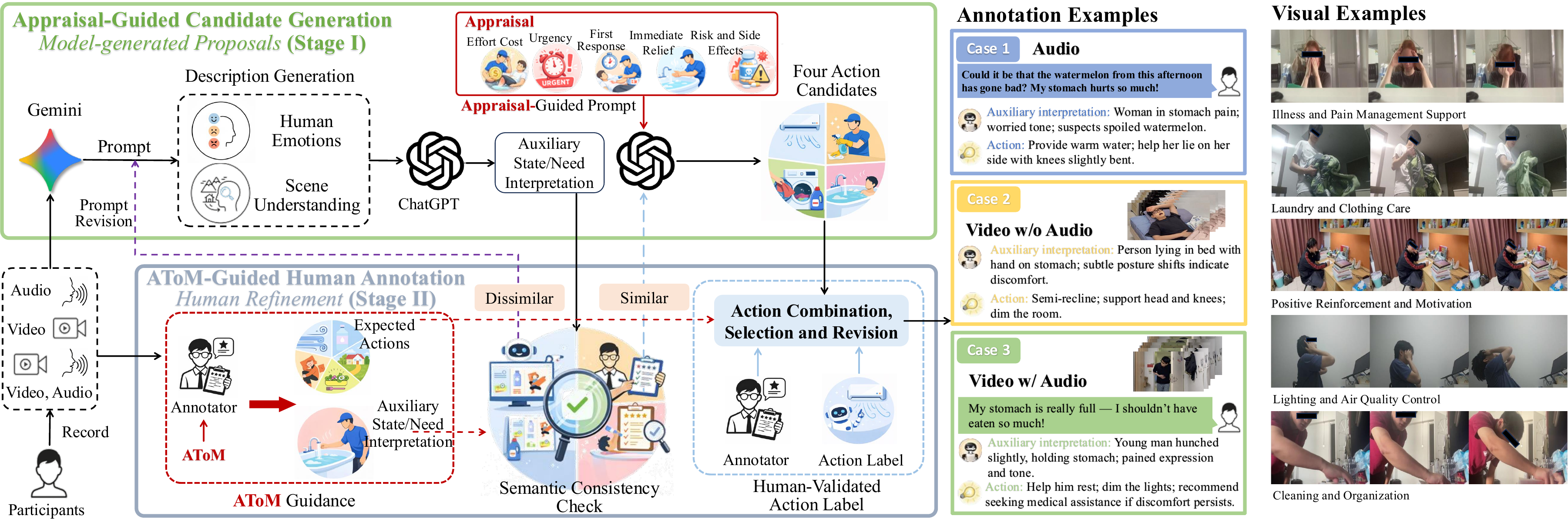}}
  \caption{\textbf{Construction of ProAction.} Stage I performs appraisal-guided candidate generation from real-world audio, video, and audio-video observations. Stage II conducts AToM-guided human annotation, where annotators inspect the original multimodal observations and validate, combine, or revise the candidate actions. Human-state/need interpretations are retained only as auxiliary semantic metadata. 
  % while the final benchmark labels are human-validated high-level robot actions.
  }\label{fig:pipeline}
  \end{center}
\vspace{-20pt}
\end{figure*}

\textbf{Overview.}
% ProAction is constructed to associate human-centered multimodal observations with contextually appropriate high-level robot actions. 
% To obtain human-aligned annotations, a two-stage human-in-the-loop annotation pipeline that combines guidance from appraisal-based criteria and Affective Theory-of-Mind reasoning is designed in Fig.~\ref{fig:pipeline}. 
ProAction is constructed to provide cognitively grounded supervision for proactive high-level action reasoning. Instead of directly assigning actions to multimodal cues, as shown in Fig.~\ref{fig:pipeline}, its two-stage human-in-the-loop pipeline explicitly incorporates appraisal-based judgment and Affective Theory-of-Mind reasoning to derive contextually appropriate robot actions.
By default, only high-level actions are used for training, while the human-state/need interpretations produced during annotation are retained as auxiliary semantic metadata.
As summarized in Tab.~\ref{compar}, existing datasets differ substantially in data grounding, modality coverage, and action abstraction. ProAction complements these resources by directly recording participants in target daily-life scenarios, synchronously capturing human-centered audio-visual observations, and pairing them with auxiliary human-state/need interpretations and human-validated high-level robot actions. This design specifically supports the upstream ProRobo problem of determining what action a robot should take before downstream execution.

\subsubsection{Dataset Collection and Participants}
ProAction is constructed from real-world human recordings designed to capture multimodal observations relevant to ProRobo. Each sample contains a 5–15s observation and is collected under one of three modality configurations: audio, video, or synchronized audio-video. These configurations are designed to capture complementary cues from speech and vocal characteristics, non-verbal human behavior, and their combination.

For \textit{audio-only} samples, participants verbally express their current situations or states in natural speech. Vocal characteristics are required to remain consistent with the corresponding scenario. For instance, discomfort may be expressed with a weak or painful tone, whereas positive events may involve excited speech. For \textit{video-only} samples, participants express the corresponding situation entirely through non-verbal cues, including body posture, facial expressions, movements, and interactions with surroundings. The recordings are captured from a third-person perspective to approximate the observation of a nearby robot, while spoken language is excluded. For \textit{audio-video} samples, verbal content and characteristics, facial expressions, body movements, and environmental interactions are synchronously recorded, allowing the modalities to provide complementary contextual evidence. All recordings are collected in indoor daily-life environments and are designed to preserve both human behavior and the surrounding context. Recording conditions are further varied across viewpoints, participant attire, and locations to increase observation diversity.

% ProAction is collected from 30 participants with backgrounds in electronics or electrical engineering, including participants from China (40\%), Singapore (33.3\%), India (20\%), and the US (6.6\%). This participant composition introduces variation in expression styles and everyday experiences. 
ProAction involves thirty participants with backgrounds in electronics or electrical engineering, including individuals from China (40\%), Singapore (33.3\%), India (20\%), and the United States (6.6\%), introducing variation in linguistic expression, behavioral styles, and everyday experiences.
Thirteen participants are further involved in the annotation process as described in Sec.~\ref{humanintheloop}. 
\textit{All participants provided informed consent for data collection and research use. This study involved only non-interventional collection of audio, video, and audio-video recordings for robotics research, without medical, psychological, or physiological intervention or manipulation, and was therefore considered exempt from formal review board approval. Personally identifiable information was anonymized before dataset release, and visual data were further processed to reduce identity exposure.}
% All participants provide informed consent for data collection and research use. Personally identifiable information is anonymized before dataset release, and visual data are further processed to reduce identity exposure.

% \begin{figure*}[t!]
%     \centering
%     \subfloat[ActPrim embeddi.]{\includegraphics[width=0.7\linewidth]{imgs/pipeline_ral.pdf}\tiny \selectfont}
%     % \vspace{-10pt}
%     \subfloat[ActPrim embeddings''.]{\includegraphics[width=0.22\linewidth]{imgs/statistics.pdf}}
%     % \vspace{-2pt}
%     \caption{\textbf{Visualization of ActPrim embeddings and corresponding visual observations with the color of borders matching the ActPrims.} Transparency decreases as the episode progresses. The ActPrim estimator with the APS learns a semantically meaningful latent space that encodes consecutive visual observations into temporally separate segments, which are human-understandable and represent the ActPrims.}
%     \label{fig:space}
%     \vspace{-10pt}
% \end{figure*}

\subsubsection{Human-in-the-Loop Annotation Pipeline}\label{humanintheloop}
High-level action annotation is inherently subjective since an appropriate response depends not only on observed scenes, but also on the person's state, urgency, contextual constraints, feasibility, and potential risks. We design a two-stage human-in-the-loop pipeline that combines scalable model-generated proposals with human judgment, as shown in Fig.~\ref{fig:pipeline}.

\textbf{Stage I: Appraisal-Guided Candidate Generation.}
Following the Component Process Model~\cite{scherer2009dynamic}, we incorporate appraisal principles into candidate action generation so that the proposed actions are both semantically relevant and practically appropriate. Specifically, five criteria are considered: \textit{effort cost} (whether the expected benefit justifies the execution effort), \textit{urgency} (whether immediate intervention is needed), \textit{human first response} (whether the action matches a reasonable first response), \textit{immediate relief} (whether it directly alleviates the observed difficulty), and \textit{risk and side effects} (whether it may introduce physical, social, or environmental risks).
% (1) \textit{Urgency}: whether the observed situation requires immediate intervention; 
% (2) \textit{Immediate relief}: whether the action can directly alleviate the observed difficulty or discomfort;
% (3) \textit{Human first response}: whether the action is consistent with how a person would reasonably respond first in a similar situation; 
% (4) \textit{Effort cost}: whether the expected benefit justifies the effort required for execution;
% (5) \textit{Risk and side effects}: whether the action may introduce physical, social, or environmental risks. 
% \vspace{-2pt}
% \begin{itemize}
%     \item[(1)] \textit{Effort cost}: whether the expected benefit justifies the effort required for execution;
%     \item[(2)] \textit{Urgency}: whether the observed situation requires immediate intervention; 
%     \item[(3)] \textit{Human first response}: whether the action is consistent with how a person would reasonably respond first in a similar situation; 
%     \item[(4)] \textit{Immediate relief}: whether the action can directly alleviate the observed difficulty or discomfort; 
%     \item[(5)] \textit{Risk and side effects}: whether the action may introduce physical, social, or environmental risks. 
% \end{itemize}
The five criteria are explicitly incorporated into the candidate-generation prompt, encouraging actions that are timely, immediately helpful, feasible, and risk-aware.
% particularly when multimodal observations are ambiguous.
For each sample, Gemini-3~\cite{gemini3} first produces a structured description of the multimodal observation, including observable human behavior, physical or emotional state, surrounding context, and salient events. Based on this description together with the original multimodal input, ChatGPT~\cite{achiam2023gpt} generates four candidate high-level robot actions under the five appraisal criteria. An auxiliary human-state/need interpretation is also retained to facilitate semantic checking during subsequent annotation.
Importantly, the interpretation does not constitute the prediction target of ProRobo, and the generated high-level actions serve as the candidate labels passed to Stage II for human review.

\textbf{Stage II: AToM-Guided Human Annotation.}
Model-generated candidates may be incomplete, overly generic, or inconsistent with subtle multimodal cues. Human annotators therefore inspect the original recordings directly rather than relying solely on model-generated scene descriptions. Affective Theory of Mind (AToM)~\cite{rao1995bdi} is used as an annotation principle to encourage consideration of the person's observable emotional and physical state, contextual constraints, and potential improvement in comfort or situation.

For each sample, annotators provide: \textit{(1) auxiliary semantic interpretation of observed human state or underlying need} and \textit{(2) the primary annotation: the expected high-level action}. 
The auxiliary interpretation makes the semantic rationale underlying the action explicit and facilitates quality control. It is retained with ProAction for interpretability, dataset analysis, and future fine-grained studies, but is not defined as an intermediate prediction target in the current benchmark.
Annotators are instructed to avoid direct cue-to-action matching and instead judge whether an action is contextually appropriate, helpful, feasible, and consistent with the observed situation.

The human-provided state interpretation is compared with the auxiliary interpretation generated in Stage I. If they are semantically consistent, the sample proceeds to final action validation. Otherwise, the model-generated interpretation is revised and the candidate actions are reconsidered. The final high-level action may be selected, combined, or rewritten by annotators when model-generated proposals are incomplete or inappropriate. 
This auxiliary semantic consistency checking reduces the influence of incomplete or hallucinated scene interpretations before final action validation.

Together, the two stages operationalize cognitive action judgment by integrating contextual appraisal and human-state reasoning before determining the final high-level action.

\begin{figure}[t]
  \centering
 \includegraphics[width=0.45\linewidth]{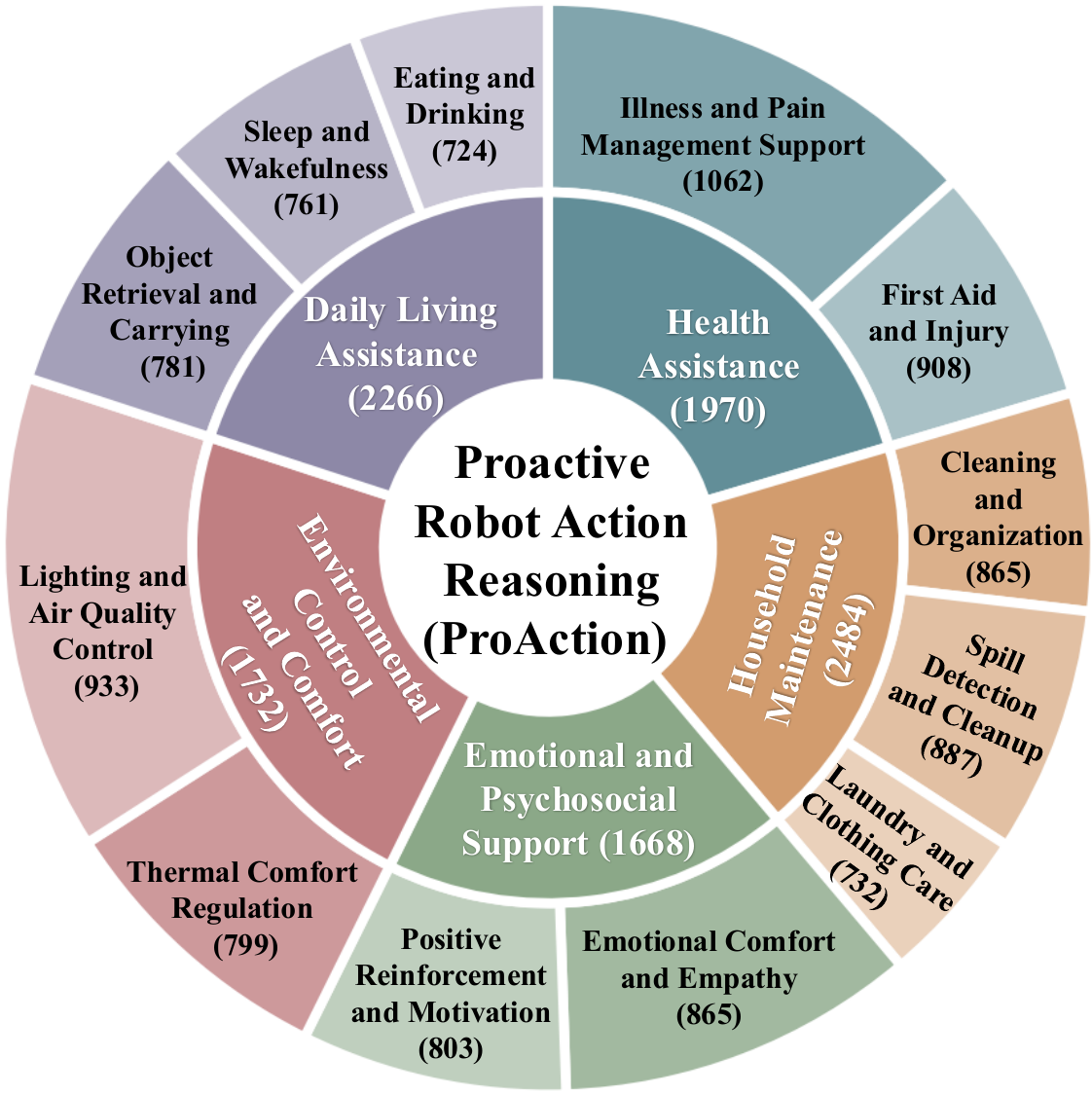}
  % \vspace{-4pt}
   \caption{\textbf{Dataset Statistics.} ProAction contains over 10K human-centered multimodal samples covering 12 daily-life scenarios across five common scenes for proactive high-level robot action reasoning.}\label{fig4}
   \vspace{-8pt}
\end{figure}

% \begin{wrapfigure}{r}{0.6\linewidth}
%     \centering
%     \vspace{-8pt}
%     \includegraphics[width=0.9\linewidth]{imgs/statistics.pdf}
%     \vspace{-2pt}
%     \caption{\textbf{Statistics of ProAction.} The dataset contains 10K multimodal samples covering 12 daily-life scenarios across five common human-centered scenes for proactive high-level robot action reasoning.}\label{fig4}
% \end{wrapfigure}

\subsubsection{Dataset Quality and Statistics}
\textbf{Data Quality Control.}
We employ multiple quality-control procedures during data collection and annotation. Each recording is checked for duration, modality consistency, contextual completeness, and consistency between the expressed human behavior and the intended scenario. Video-only recordings are additionally checked to ensure that complete human behavior and environmental context are visible without recording artifacts. Audio and audio-video samples are checked for consistency among speech content, vocal characteristics, observed behavior, and scenario context. Annotators further self-check their labels for grammatical or typographical errors and cross-check annotations produced by other annotators. Identified errors are corrected before the sample is included in the final dataset.

\begin{table*}[t]
\caption{\textbf{Performance (\%) of representative MLLMs and MMC2Act on ProAction}. Results are reported on both the in-distribution (ID) and subject-disjoint out-of-distribution (OOD) test sets. ``A'', ``V'', and ``T'' denote audio, video, and text, respectively.}\label{Table2}
% \vspace{-5pt}
	\centering
    \setlength{\tabcolsep}{4pt}
    \resizebox{0.6\linewidth}{!}{
		\begin{tabular}{llccccccccccccccccc}
\toprule
\multirow{2}{*}{\textbf{Setting}} & \multirow{2}{*}{\textbf{Model}} & \multirow{2}{*}{\textbf{\#Params}} & \multirow{2}{*}{\textbf{Open}} & \multicolumn{4}{c}{\textbf{In-Distribution Split}} & \multicolumn{4}{c}{\textbf{Out-of-Distribution Split}} \\
\cmidrule(lr){5-8} \cmidrule(lr){9-12}
& & & \textbf{Source} & Overlap & LCS & AvgSim. & ST-Sim. & Overlap & LCS & AvgSim. & ST-Sim. \\
\midrule
\multirow{6}{*}{A+T} 
& GLM-4~\cite{glm2024chatglm} & 9B & \textcolor{green!70!black}{\ding{52}} & 36.9 & 32.4 & 49.9 & 40.2 & 29.5 & 18.4 & 40.3 & 40.8 \\
& Llama-Omni~\cite{fang2025llama} & 8B & \textcolor{green!70!black}{\ding{52}} & 33.4 & 28.4 & 28.6 & 32.4 & 29.2 & 18.3 & 44.1 & 46.8 \\
& Gemini-3~\cite{gemini3} & - & \textcolor{purple}{\XSolidBrush} & 37.8 & 28.4 & 35.2 & 38.0 & 31.8 & 20.6 & 45.1 & 43.1 \\
& Qwen2.5-Omni~\cite{xu2025qwen2} & 3B & \textcolor{green!70!black}{\ding{52}} & 47.1 & 38.2 & 50.8 & 51.1 & 27.4 & 17.5 & 45.7 & 42.2  \\
& Qwen2.5-Omni~\cite{xu2025qwen2} & 7B & \textcolor{green!70!black}{\ding{52}} & 43.4 & 35.3 & 54.3 & 52.0 & 26.0 & 16.5 & 42.7 & 42.4  \\
& \textbf{MMC2Act (Ours)} & 7B & \textcolor{green!70!black}{\ding{52}} & \textbf{51.4} \textcolor{green!70!black}{$\uparrow$} & \textbf{43.4} \textcolor{green!70!black}{$\uparrow$} & \textbf{60.6} \textcolor{green!70!black}{$\uparrow$} & \textbf{58.4} \textcolor{green!70!black}{$\uparrow$} &  \textbf{48.5} \textcolor{green!70!black}{$\uparrow$} & \textbf{36.9} \textcolor{green!70!black}{$\uparrow$} & \textbf{49.9} \textcolor{green!70!black}{$\uparrow$} & \textbf{45.0} \textcolor{green!70!black}{$\uparrow$} \\
% & \textbf{Need2Act (Ours)} & 7B & \textcolor{green!70!black}{\ding{52}} & \textbf{55.0} & \textbf{47.5} & \textbf{60.2} & \textbf{59.4} & \textbf{75.0} & \textbf{59.4} \\
\midrule
\multirow{12}{*}{V+T} 
& ChatGPT-4o~\cite{chatgpt4o} & - & \textcolor{purple}{\XSolidBrush} & 27.9 & 22.7 & 36.3 & 40.6 & 20.5 & 13.2 & 33.8 & 34.3 \\
& Gemini-3~\cite{gemini3} & - & \textcolor{purple}{\XSolidBrush} & 35.0 & 29.0 & 37.2 & 41.5 & 24.2 & 15.6 & 41.1 & 34.1 \\
& Gemini-2.5~\cite{comanici2025gemini} & - & \textcolor{purple}{\XSolidBrush} & 32.0 & 29.3 & 34.6 & 36.5 & 26.3 & 17.1 & 37.7 & 40.2 \\
% & InternVL2.5~\cite{chen2024expanding} & 1B & \textcolor{green!70!black}{\ding{52}} & 17.9 & 13.9 & 28.5 & 22.1 &  15.2 & 10.5 & 30.4 & 10.9 \\
% & InternVL2.5~\cite{chen2024expanding} & 2B & \textcolor{green!70!black}{\ding{52}} & 20.9 & 15.5 & 27.6 & 29.4 &  11.8 & 7.3 & 19.1 & 9.4 \\
& InternVL2.5~\cite{chen2024expanding} & 8B & \textcolor{green!70!black}{\ding{52}} & 33.3 & 25.2 & 34.2 & 37.2 &  15.2 & 9.3 & 27.9 & 20.3  \\
% & InternVL3~\cite{zhu2025internvl3} & 1B & \textcolor{green!70!black}{\ding{52}} & 14.6 & 11.8 & 27.9 & 20.3 &  &  \\
& LLaVA-Video~\cite{zhang2025llavavideo} & 7B & \textcolor{green!70!black}{\ding{52}} & 14.6 & 11.8 & 27.9 & 20.3 & 17.9 & 13.9 & 28.5 & 22.1 \\
& InternVL3~\cite{zhu2025internvl3} & 2B & \textcolor{green!70!black}{\ding{52}} & 20.9 & 15.3 & 32.3 & 30.8 &  13.3 & 8.7 & 25.7 & 15.6 \\
& InternVL3~\cite{zhu2025internvl3} & 8B & \textcolor{green!70!black}{\ding{52}} & 19.2 & 15.1 & 31.2 & 32.8 &  14.2 & 9.0 & 28.3 & 16.6 \\
& InternVL3~\cite{zhu2025internvl3} & 9B & \textcolor{green!70!black}{\ding{52}} & 23.5 & 17.9 & 20.2 & 31.2 &  16.6 & 10.9 & 29.0 & 17.1 \\
& Video-Llama~\cite{zhang2023video} & 7B & \textcolor{green!70!black}{\ding{52}} & 17.3 & 14.0 & 28.1 & 23.1 & 29.1 & 16.6 & 14.4 & 17.0 \\
& Qwen2.5-Omni~\cite{xu2025qwen2} & 3B & \textcolor{green!70!black}{\ding{52}} & 24.9 & 19.9 & 31.6 & 25.8 & 19.8 & 12.0 & 34.6 & 25.5 \\
& Qwen2.5-Omni~\cite{xu2025qwen2} & 7B & \textcolor{green!70!black}{\ding{52}} & 25.9 & 20.6 & 33.7 & 33.1 & 19.0 & 12.0 & 30.8 & 32.4 \\
& \textbf{MMC2Act (Ours)} & 7B & \textcolor{green!70!black}{\ding{52}} & \textbf{46.9} \textcolor{green!70!black}{$\uparrow$} & \textbf{38.5} \textcolor{green!70!black}{$\uparrow$} & \textbf{48.2} \textcolor{green!70!black}{$\uparrow$} & \textbf{45.3} \textcolor{green!70!black}{$\uparrow$} & \textbf{46.9} \textcolor{green!70!black}{$\uparrow$} & \textbf{36.3} \textcolor{green!70!black}{$\uparrow$} & \textbf{44.7} \textcolor{green!70!black}{$\uparrow$} & \textbf{39.4} \textcolor{green!70!black}{$\uparrow$} \\
% & \textbf{Need2Act (Ours)} & 7B & \textcolor{green!70!black}{\ding{52}} & \textbf{47.0} & \textbf{38.2} & \textbf{45.3} & \textbf{45.3} & \textbf{74.1} & \textbf{55.0} \\
\midrule
\multirow{6}{*}{A+V+T} 
& Gemini-2.5~\cite{comanici2025gemini} & - & \textcolor{purple}{\XSolidBrush} & 31.2 & 26.5 & 32.6 & 37.6 & 27.2 & 17.6 & 37.6 & 39.7  \\
& Gemini-3~\cite{gemini3} & - & \textcolor{purple}{\XSolidBrush} & 33.1 & 35.4 & 37.5 & 40.5 & 31.8 & 20.9 & 46.8 & 41.6 \\
& Ming-Omni~\cite{ai2025ming} & 3B & \textcolor{green!70!black}{\ding{52}} & 31.4 & 23.8 & 43.5 & 30.8 & 28.0 & 17.9 & 44.2 & 40.0 \\
& Video-Llama2~\cite{cheng2024videollama} & 7B & \textcolor{green!70!black}{\ding{52}} & 24.9 & 19.2 & 21.1 & 22.0 & 22.5 & 14.8 & 14.2 & 15.0 \\
% & QWen2.5-Omni~\cite{xu2025qwen2} & 3B & \textcolor{green!70!black}{\ding{52}} & 31.4 & 23.8 & 43.5 & 30.8 & 28.0 & 17.9 & 44.2 & 40.0 \\
& Qwen2.5-Omni~\cite{xu2025qwen2} & 7B & \textcolor{green!70!black}{\ding{52}} & 36.0 & 28.2 & 42.4 & 43.0 & 27.5 & 17.4 & 42.8 & 41.7 \\
% QWen2.5-Omni
& \textbf{MMC2Act (Ours)} & 7B & \textcolor{green!70!black}{\ding{52}} & \textbf{54.4} \textcolor{green!70!black}{$\uparrow$} & \textbf{47.4} \textcolor{green!70!black}{$\uparrow$} & \textbf{69.3} \textcolor{green!70!black}{$\uparrow$} & \textbf{67.8} \textcolor{green!70!black}{$\uparrow$} & \textbf{50.7} \textcolor{green!70!black}{$\uparrow$} & \textbf{37.9} \textcolor{green!70!black}{$\uparrow$} & \textbf{54.3} \textcolor{green!70!black}{$\uparrow$} & \textbf{50.1} \textcolor{green!70!black}{$\uparrow$} \\
% & \textbf{Need2Act (Ours)} & 7B & \textcolor{green!70!black}{\ding{52}} & \textbf{52.1} & \textbf{45.6} & \textbf{57.6} & \textbf{54.5} & \textbf{76.9} & \textbf{57.4} \\
% & \textbf{MNU-MLLM} & 3B & Open & \textbf{60.5} & \textbf{52.2} & \textbf{62.7} & \textbf{61.2} & \textbf{78.0} & \textbf{62.9} \\
\bottomrule
\end{tabular}}
\vspace{-5pt}
\end{table*}

\textbf{Annotation Reliability.}
We conduct an independent inter-annotator agreement study on 200 samples stratified across different scenarios and scenes. Three annotators independently relabel the high-level robot action without access to either the original ground-truth annotation or model-generated candidates. Since the annotations are open-ended natural-language descriptions rather than discrete categories, conventional agreement metrics such as Fleiss' $\kappa$~\cite{Fleiss} are not directly applicable. We therefore encode the annotations using Sentence-BERT and compute semantic agreement. The average pairwise cosine similarity reaches \textit{0.82}, while \textit{98.6\%} of samples achieve majority semantic agreement at a similarity threshold of 0.75. Similar agreement is observed when the threshold varies from 0.70 to 0.80, supporting the reliability of the high-level robot-action annotations.

\textbf{Human Intervention in Annotation.}
To quantify the influence of model-generated proposals, we measure the edit distance between generated candidates and final human-authored actions. 
% The average character edit rate is 58.21\%, indicating substantial rewriting rather than superficial modification. Moreover, 41.25\% of samples require additional adjudication during annotation. 
Annotators revise 98.17\% of model-generated proposals, with an average character edit rate of 58.21\%, indicating substantial rewriting. Moreover, 41.25\% of samples further require adjudication. These results show that the final labels are not directly inherited from model-generated candidates. Model outputs serve only as initial proposals, and the final action annotations are determined through human validation.

\textbf{Dataset Statistics.}
ProAction contains 12 daily-life scenarios across five scenes: health assistance, household maintenance, emotional and psychosocial support, environmental control and comfort, and daily living assistance, as summarized in Fig.~\ref{fig4}. They cover both physically grounded assistance and socially situated responses, ranging from object retrieval and spill cleanup to emotional comfort and positive reinforcement, and contain a total of around 10K samples. 
\vspace{-5pt}

\subsection{MMC2Act: A Validation Baseline for ProAction}
% To examine whether ProAction is effective for ProRobo, we develop \textit{MMC2Act}, a validation baseline to infer high-level actions from multimodal cues. MMC2Act uses the auxiliary human-state/need annotations only as additional supervision during training. The primary prediction target remains the high-level robot action.
% We implement MMC2Act, a Perceiver--Reasoner--Speaker validation baseline for high-level action reasoning. For physical deployment, its inferred actions are subsequently mapped to available robot behaviors through a separate execution interface, as illustrated in Fig.~\ref{framework} (b).
% We implement MMC2Act, a Perceiver--Reasoner--Speaker validation baseline for high-level action reasoning, 
We implement MMC2Act, a validation baseline built upon Qwen2.5-Omni-7B with a \textit{Perceiver--Reasoner--Speaker} structure for high-level action reasoning, as illustrated in Fig.~\ref{framework} (b). For physical deployment in Fig.~\ref{framework} (c), the inferred high-level actions are grounded to a predefined set of executable robot skills through a lightweight rule-based interface.
% For physical deployment, its inferred actions are subsequently mapped to available robot behaviors through a separate execution interface (see Fig.~\ref{framework} (c)).

\textbf{Multimodal Perceiver.}
Given visual observation $V$, audio signal $A$, and textual input $T$, the native multimodal encoders, including $f_v$ for vision, $f_a$ for audio, and $f_t$ for text, respectively produce modality-specific representations:
{\setlength\abovedisplayskip{3pt}
\setlength\belowdisplayskip{3pt}
\begin{equation}
X=[v;a;c], \quad v=f_v(V),\; a=f_a(A),\; c=f_t(T).
\end{equation}}The embedding $v$, $a$, and $c$ integrate complementary information from different modalities, while $X$ provides a unified multimodal representation for downstream action reasoning.
% The resulting embedding $v$, $a$, and $c$ integrate complementary information from human behavior, surrounding context, speech content, vocal characteristics, and textual context. The multimodal representation $X$ provides a unified multimodal representation for downstream high-level action reasoning.

% \textbf{Perceiver for Need Perception.}
% The Perceiver encodes heterogeneous inputs into a unified embedding $\mathbf{X}$, following the multimodal processing pipeline of Qwen2.5-Omni \cite{xu2025qwen2}:
% {\setlength\abovedisplayskip{2pt}
% \setlength\belowdisplayskip{2pt}
% \begin{equation}
%     \mathbf{X}=[\mathbf{v};\mathbf{a};\mathbf{c}], \ \text{where} \ \ \mathbf{v}=f_v(I), \ \mathbf{a}=f_a(A), \ \mathbf{c}=f_t(C).
% \end{equation}}$\mathbf{v}$ captures body posture and facial expression, % and surrounding context
% while $\mathbf{a}$ captures speech content and vocal tone. Finally, $\mathbf{X}$ encodes latent human needs
% % embedded in $I$, $A$, and $C$ 
% and serves as input to the Reasoner for high-level action reasoning.

\textbf{Action Reasoner.}
The multimodal representation $X$ is directly used to generate a high-level action. The reasoner autoregressively generates text
tokens $Y=(y_1,\ldots,y_L)$:
{\setlength\abovedisplayskip{3pt}
\setlength\belowdisplayskip{3pt}
\begin{equation}
p_{\theta}(Y|X)={\textstyle \prod_{\ell=1}^{L}}p_{\theta}(y_{\ell}\mid X,y_{<\ell}).
\end{equation}}The training objective is the causal language-modeling loss over the high-level action annotation:
{\setlength\abovedisplayskip{3pt}
\setlength\belowdisplayskip{3pt}
\begin{equation}
\mathcal{L}_{\mathrm{act}}
= {\textstyle -\sum_{\ell=1}^{L}} \log p_{\theta} (\bar{y}_{\ell}\mid X,\bar{y}_{<\ell}).
\end{equation}}MMC2Act is trained on high-level actions without auxiliary human-state/need supervision.
Thus, MMC2Act does not implement an explicit cognitive reasoning chain. Instead, it implicitly learns observation-to-action mapping encoded in the cognitively constructed high-level action supervision.

\textbf{Speaker and Action Execution.}
The inferred action tokens are converted to speech through a Speaker module for human--robot interaction. 
% For robot execution, the grounded skill is executed through the action-execution interface in K1, which is not learned and is used only for physical grounding.
For robot execution, the inferred high-level action is grounded to a predefined executable skill through the same rule-based K1 interface, which is not learned and is used only for physical grounding.
For potentially sensitive actions involving physical contact or medical assistance, the system first requests user confirmation before execution.
% \vspace{-5pt}

\begin{figure*}[t!]
  \begin{center}
\centerline{\includegraphics[width=0.95\linewidth]{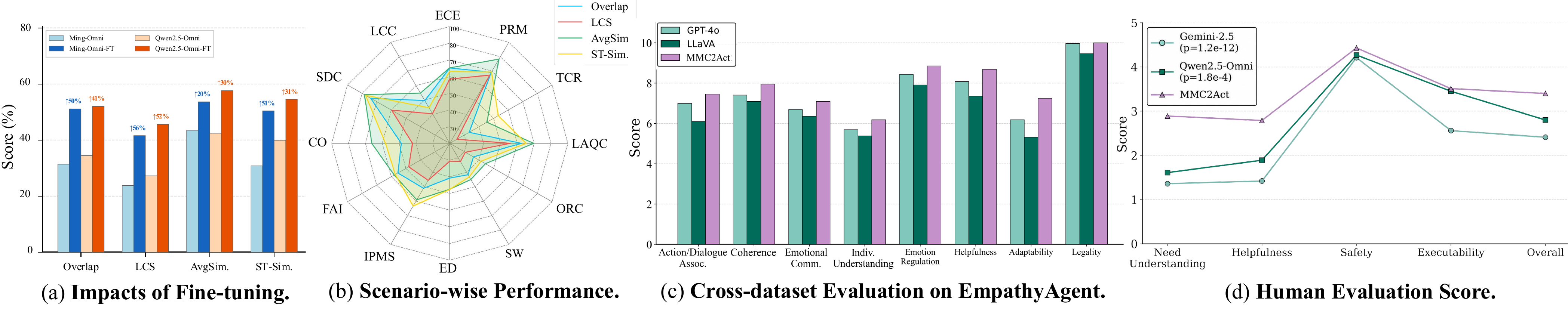}}
\vspace{-5pt}
  \caption{\textbf{More Analysis.} (a) Effect of ProAction fine-tuning (FT) on two models. (b) Scenario-wise performance across 12 scenarios. (c) Cross-dataset transfer evaluation on EmpathyAgent. (d) Human evaluation on action quality. The abbreviations in (b) denote different scenarios.}\label{scenarioCcross}
  \end{center}
  \vspace{-20pt}
\end{figure*}

\section{Experiments}\label{sec:exp}
We evaluate ProAction through five research questions (RQ1--RQ5) on task difficulty, supervision, multimodal cues, generalization, and physical action grounding.

\subsection{Experimental Setup}\label{sec:metrics}
\subsubsection{Data Split}
We divide ProAction into training, validation, in-distribution (ID) test, and subject-disjoint test sets with ratios of 80\%, 10\%, 5\%, and 5\%, respectively. The ID test set follows the participant distribution of the training data, whereas the subject-disjoint test set contains participants who do not appear during training or validation.
This setting allows us to separately evaluate action reasoning under in-domain distributions and generalization to unseen human appearances, voices, speaking styles, and behavioral patterns. In all settings, $T$ denotes the same task-level textual prompt, while the available sensory modalities vary.

\subsubsection{Evaluation Metrics}
Since ProRobo requires high-level action generation rather than classification over a predefined action set, we evaluate generated actions against human-validated reference actions using both lexical and semantic metrics.
We report (1) \textit{Overlap}, which measures token-level overlap between generated and reference actions; (2) \textit{LCS}, which measures their longest common subsequence; (3) \textit{AvgSim.}, which measures semantic similarity to the corresponding cluster of human action responses;
% computes the average semantic similarity between generated and reference action descriptions; 
and (4) \textit{ST-Sim.}, which measures sentence-level semantic similarity using sentence embeddings.
Together, these metrics capture both surface-level correspondence and semantic consistency between predicted and human-validated high-level actions.

\subsubsection{Baselines}
We evaluate representative general-purpose MLLMs to examine their intrinsic capability for ProRobo. We select two popular closed-source MLLMs, \textit{i.e.,} Gemini~\cite{comanici2025gemini,gemini3} and ChatGPT~\cite{chatgpt4o}. We further include representative open-source models from InternVL \cite{chen2024expanding,zhu2025internvl3}, Video-Llama \cite{zhang2023video}, Video-Llama2~\cite{cheng2024videollama}, Llama-Omni~\cite{fang2025llama}, LLaVA-Video \cite{zhang2025llavavideo}, GLM \cite{glm2024chatglm}, Ming-Omni~\cite{ai2025ming} and Qwen2.5-Omni \cite{xu2025qwen2}. Methods based on image-text inputs \cite{chen2023shikra,peng2023kosmos} or adopting VQA as an auxiliary task \cite{yang2025magma} are excluded. By default, all models are evaluated on high-level action generation.

\subsubsection{Implementation Details}
For MMC2Act, we fine-tune Qwen2.5-Omni-7B with LoRA \cite{hu2022lora} while freezing encoders for 4 epochs on 4 A100 GPUs, with a learning rate of $5\times10^{-5}$ and a batch size of 2. A weight decay of $0.1$ is applied for regularization, and the Adam parameter $\beta$ is set to $(0.9, 0.95)$. We follow the multimodal processing pipeline in Qwen2.5-Omni to optimize the likelihood of generating appropriate text responses. For real-world deployment, we adopt a 22-DoF Booster K1 robot \cite{booster_k1}, equipped with a binocular camera, a microphone array, a speaker, and a 9-axis IMU. Eight scenarios are tested: ``Thirsty'', ``Cold'', ``Messy'', ``Sad'', ``Excited'', ``Spill Water'', ``Dark'' and ``Sleepy''.
A lightweight safety-aware grounding layer is applied: actions involving physical contact, medical intervention, ambiguity, low confidence, or unavailable capabilities trigger confirmation, clarification, verbal assistance, or abstention rather than direct execution.
\vspace{-5pt}

\subsection{RQ1: Is ProRobo Challenging?}\label{question1}
We first evaluate representative general-purpose MLLMs on ProAction under the zero-shot setting to examine whether their multimodal reasoning capabilities are sufficient for ProRobo. The results are shown in Tab.~\ref{Table2}. It is observed that general-purpose MLLMs exhibit limited performance in inferring appropriate high-level actions from human-centered multimodal cues. Although these models possess strong general visual-language or audio-visual reasoning capabilities, their generated responses frequently capture only salient scene content or human behavior without producing sufficiently specific and executable actions. This suggests that general-purpose multimodal reasoning alone is insufficient for ProRobo, and task-specific action supervision is needed to better align multimodal understanding with appropriate high-level robot actions.
% \vspace{-5pt}

\begin{figure*}[t!]
  \begin{center}
  \centerline{\includegraphics[width=0.92\linewidth]{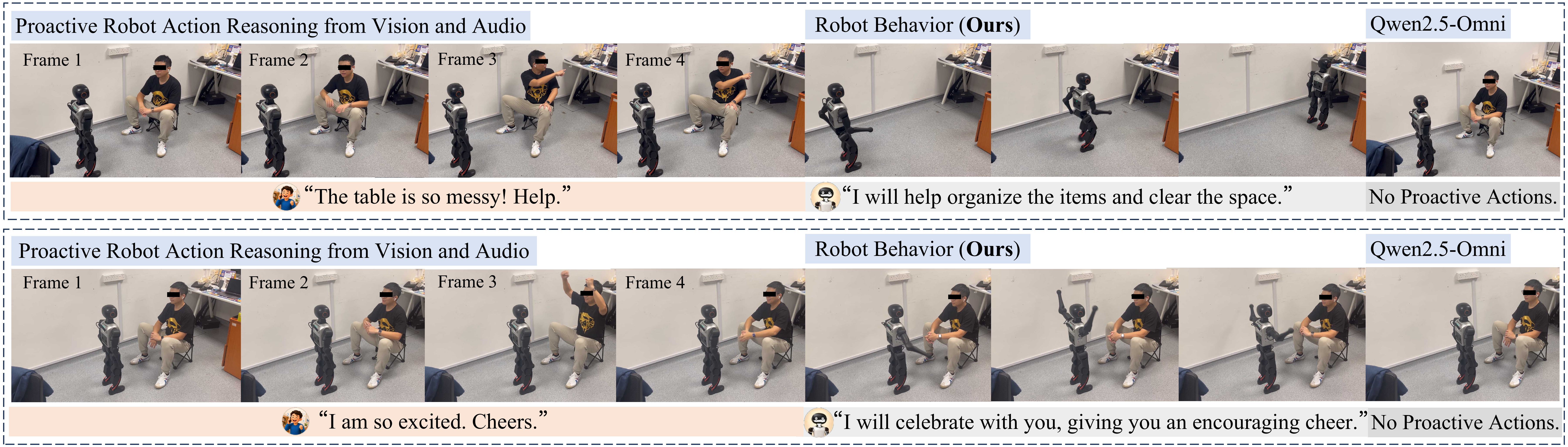}}
  \caption{\textbf{Physical robot deployment.} Given multimodal human-centered observations, MMC2Act infers an appropriate high-level action, which is further grounded into a predefined executable behavior on a Booster K1 robot. The examples show table organization and an encouraging response, while Qwen2.5-Omni produces no proactive action under the same observations.}\label{fig:fig5}
  \end{center}
  \vspace{-20pt}
\end{figure*}

\subsection{RQ2: Does ProAction Provide Effective Supervision for Cognitive Action Reasoning?}\label{sec:modality}
\subsubsection{ProAction-based Fine-tuning} We next examine whether ProAction provides effective supervision for ProRobo by fine-tuning two different backbones, Ming-Omni-3B~\cite{ai2025ming} and Qwen2.5-Omni-3B~\cite{xu2025qwen2}, on the same action annotations, comparing them with their corresponding zero-shot counterparts. As shown in Fig.~\ref{scenarioCcross} (a), fine-tuning consistently improves performance across both lexical and semantic metrics, demonstrating that ProAction provides useful supervision. Similar results are also observed for both Ming-Omni and Qwen2.5-Omni. The consistent gains across two distinct backbones suggest that the benefit of ProAction supervision is not backbone-specific. These results indicate that models can effectively learn the observation-to-action mapping encoded in ProAction's cognitively grounded high-level action supervision.

\subsubsection{Utilization of Human-State/Need Annotations} We further evaluate auxiliary human-state/need supervision by additionally training the action reasoner to generate the corresponding state/need interpretation. This yields an average improvement of only 1.53\% across the four ID metrics, suggesting their complementary semantic value but limited benefit under the current strategy. Thus, more effective methods are needed to fully exploit these annotations.
% \vspace{-5pt}

\subsection{RQ3: How Do Different Cues Contribute to ProRobo?}
A central motivation of ProAction is that appropriate robot actions may depend on complementary information distributed across human-centered modalities. For modality comparison, audio- and video-only results are obtained from the same synchronized audio-video samples by masking the other modality. As shown in Tab.~\ref{Table2}, both audio and visual inputs provide useful information for high-level action reasoning. For MMC2Act, audio consistently outperforms vision, likely because speech semantics, vocal characteristics, and nonverbal sounds provide more direct cues of human conditions, whereas vision captures body posture, facial expressions, object interactions, and environmental context. Combining both modalities achieves the best performance, confirming that audio and vision provide complementary rather than redundant evidence. In contrast, general-purpose MLLMs do not always benefit consistently from additional modalities, suggesting that effective multimodal integration is important. These results further support the synchronized audio-visual design of ProAction.
% \vspace{-5pt}

\subsection{RQ4: How Robust Is ProAction Across Participants, Scenarios, and Datasets?}\label{sec:generalization}
\subsubsection{Subject-Disjoint Generalization}
We first evaluate the subject-disjoint out-of-distribution generalization on ProAction, where subjects are not observed during training.
As shown in Tab.~\ref{Table2}, performance decreases,
% relative to unseen subjects, 
indicating that proactive action reasoning remains sensitive to person-specific multimodal characteristics.
Nevertheless, MMC2Act retains meaningful action-generation capability on unseen subjects, suggesting that ProAction supervision provides transferable cues that go beyond memorizing individual participants.

\subsubsection{Scenario-Wise Robustness}
We further analyze performance across the 12 scenarios, where the difficulty varies substantially. As shown in Fig.~\ref{scenarioCcross} (b), situations with salient multimodal evidence generally yield stronger performance, whereas scenarios requiring more contextual interpretation or involving multiple plausible actions remain challenging. Importantly, the large variation across scenarios indicates that ProRobo cannot be reduced to simply identifying a broad scene category and retrieving a fixed action template. Instead, appropriate actions depend on human-specific behavior, multimodal cues, and environmental context within each observation.

\subsubsection{Cross-Dataset Transfer}
To examine whether the improvement on ProAction is dataset-specific or not, we evaluate MMC2Act on EmpathyAgent~\cite{chen2025empathyagent}. Fig.~\ref{scenarioCcross} (c) reports the results. It is observed that MMC2Act exhibits competitive transfer performance on several action-quality dimensions, suggesting that the learned supervision can transfer beyond ProAction.

Together, these results indicate that ProAction supervision transfers beyond memorizing individual participants or fixed scenarios, although performance remains sensitive to ambiguous and person-specific cues.
\subsection{RQ5: Are the Inferred Actions Human-Appropriate and Physically Executable?}
\subsubsection{Human Appropriateness}
Metrics in Tab.~\ref{Table2} cannot fully capture whether generated actions are appropriate for real-world interaction. We therefore randomly sample 500 test examples covering all scenarios and conduct a blind human evaluation of anonymized outputs from MMC2Act, Qwen2.5-Omni, and Gemini-2.5. Five evaluators rate each generated action on a 1--5 scale across five dimensions. MMC2Act achieves the highest scores across all dimensions, as shown in Fig.~\ref{scenarioCcross} (d). The gains are particularly evident in Need Understanding and Helpfulness, while all models receive relatively high Safety scores. The overall improvements over Qwen2.5-Omni ($p=1.8\times10^{-4}$) and Gemini-2.5 ($p=1.2\times10^{-12}$) are statistically significant under paired bootstrap tests, indicating better alignment with human judgments.

% \textbf{Key Insights.}
% First, general-purpose MLLMs remain limited in inferring appropriate actions from multimodal cues. Second, fine-tuning on ProAction substantially improves high-level action reasoning, demonstrating that the dataset provides learnable supervision beyond generic multimodal capabilities. Third, audio and vision provide complementary information, while the remaining subject- and scenario-level generalization gaps highlight the need for stronger multimodal understanding rather than simple scenario memorization.

% \subsection{Physical Robot Deployment}\label{sec:robot}
\subsubsection{Physical Executability}\label{real-expres}
We further examine whether inferred actions can be grounded into executable behaviors on a Booster K1 robot. Fig.~\ref{fig:fig5} gives two examples. In the ``Excited'' scenario, K1 first perceives multimodal cues by visual and auditory sensing. MMC2Act then infers the high-level action “Celebrate with him.”, generates a verbal response, and finally executes the corresponding “wave” behavior.
% then reasons a high-level action -- ``Celebrate with him.'' by MMC2Act, gives a verbal response, and finally executes the action ``wave''. 
The average response time is 3.1 seconds on an RTX 5090 GPU. We also test Qwen2.5-Omni and observe that clarifications are asked and no proactive actions are inferred. These results demonstrate the feasibility of grounding ProAction-trained high-level action reasoning on a physical robot. 

\subsubsection{Failure Analysis}
We observe several failure modes.
First, \textit{ambiguous multimodal cues} may support multiple plausible actions, causing the model to commit to an action despite insufficient evidence.
Second, \textit{cross-modal inconsistency} may cause the model to over-rely on one modality when visual and auditory cues provide conflicting or incomplete evidence.
Third, \textit{unsafe actions} may arise when semantically plausible predictions are inappropriate for physical execution, revealing a gap between contextual relevance and real-world safety.

\section{Conclusion and Future Work}
% This paper studied \textit{Proactive Robot Action Reasoning} (\textit{ProRobo}) as an upstream cognitive decision problem that enables robots to determine appropriate high-level actions from human-centered multimodal observations before downstream execution.
% To support it, we introduced \textit{ProAction}, a real-world multimodal dataset with 10K human-centered samples and human-aligned action annotations. Evaluation showed that ProAction provided effective supervision for ProRobo, while physical-robot deployment further validated that the inferred actions can be grounded into executable robot behaviors. 
We studied \textit{Proactive Robot Action Reasoning} (\textit{ProRobo}) as a cognitive high-level decision problem in which robots determine appropriate actions from human-centered multimodal observations. \textit{ProAction} provides cognitively grounded supervision by explicitly incorporating appraisal-based judgment and Affective Theory-of-Mind reasoning into the construction of human-aligned high-level action labels. Models trained on these labels implicitly learn the resulting observation-to-action mapping without requiring an explicit cognitive architecture. Experiments demonstrate the effectiveness and generalization of this supervision, while physical-robot deployment further validates that the inferred high-level actions can be grounded into executable robot behaviors.

A promising direction for future work is to improve the generality and reliability of ProRobo in more complex scenarios. The current study does not fully cover long-tail, personalized, or ambiguous multimodal situations. The auxiliary human-state/need annotations may support finer-grained reasoning, while observed failures motivate uncertainty-aware decision mechanisms. Future work will extend ProAction to more diverse situations and integrate ProRobo with closed-loop policies to determine not only what action to take, but also whether and when to act safely. 

We hope ProAction will advance cognitively grounded, human-centered proactive robot decision-making.
% We hope ProAction will provide a foundation for cognitively grounded, human-centered proactive robot decision-making.

% \section*{Acknowledgments}
% This should be a simple paragraph before the References to thank those individuals and institutions who have supported your work on this article.
 {
    \small
    \bibliographystyle{IEEEtran}
    \bibliography{example}
}

% \bibliographystyle{IEEEtran}

% \bibitem{ref1}
% {\it{Mathematics Into Type}}. American Mathematical Society. [Online]. Available: https://www.ams.org/arc/styleguide/mit-2.pdf

% \bibitem{ref2}
% T. W. Chaundy, P. R. Barrett and C. Batey, {\it{The Printing of Mathematics}}. London, U.K., Oxford Univ. Press, 1954.

% \bibitem{ref3}
% F. Mittelbach and M. Goossens, {\it{The \LaTeX Companion}}, 2nd ed. Boston, MA, USA: Pearson, 2004.

% \bibitem{ref4}
% G. Gr\"atzer, {\it{More Math Into LaTeX}}, New York, NY, USA: Springer, 2007.

% \bibitem{ref5}M. Letourneau and J. W. Sharp, {\it{AMS-StyleGuide-online.pdf,}} American Mathematical Society, Providence, RI, USA, [Online]. Available: http://www.ams.org/arc/styleguide/index.html

% \bibitem{ref6}
% H. Sira-Ramirez, ``On the sliding mode control of nonlinear systems,'' \textit{Syst. Control Lett.}, vol. 19, pp. 303--312, 1992.

% \bibitem{ref7}
% A. Levant, ``Exact differentiation of signals with unbounded higher derivatives,''  in \textit{Proc. 45th IEEE Conf. Decis.
% Control}, San Diego, CA, USA, 2006, pp. 5585--5590. DOI: 10.1109/CDC.2006.377165.

% \bibitem{ref8}
% M. Fliess, C. Join, and H. Sira-Ramirez, ``Non-linear estimation is easy,'' \textit{Int. J. Model., Ident. Control}, vol. 4, no. 1, pp. 12--27, 2008.

% \bibitem{ref9}
% R. Ortega, A. Astolfi, G. Bastin, and H. Rodriguez, ``Stabilization of food-chain systems using a port-controlled Hamiltonian description,'' in \textit{Proc. Amer. Control Conf.}, Chicago, IL, USA,
% 2000, pp. 2245--2249.

% \end{thebibliography}

\newpage

\vfill

\end{document}